\documentclass[11pt,a4paper]{article}
\usepackage[nohyperref]{emnlp2018}
\usepackage{times}
\usepackage{latexsym}
\usepackage{url}
\usepackage{times}
\usepackage{latexsym}
\usepackage{pgf}
\usepackage{tikz}
\usepackage{amsmath}
\usepackage{booktabs}
\usepackage{CJKutf8}
\aclfinalcopy
\newcommand{\themainpapercontent}[1]{#1}
\newcommand{\thesupplementarycontent}[1]{}

\newcommand{\hichar}[1]{\textcolor{darkred}{#1}}
\newcommand{\out}[1]{}

\colorlet{darkred}{red!50!black}
\usetikzlibrary{decorations,arrows,shapes}

\title{State-of-the-art Chinese Word Segmentation with Bi-LSTMs}

\author{Ji Ma \hspace{0.2in} Kuzman Ganchev \hspace{0.2in} David Weiss \\
 Google AI Language \\
 {\{\tt maji,kuzman,djweiss\}@google.com} }

\date{}

\begin{document}
\maketitle

\themainpapercontent{

\begin{abstract}
  A wide variety of neural-network architectures have been proposed for the task of Chinese word segmentation.  
  Surprisingly, we find that a bidirectional LSTM model, when combined with standard deep learning techniques and best practices, can achieve better accuracy on many of the popular datasets as compared to models based on more complex neural-network architectures.
  Furthermore, our error analysis shows that out-of-vocabulary words remain challenging for neural-network models, and many of the remaining errors are unlikely to be fixed through architecture changes.
  Instead, more effort should be made on exploring resources for further improvement.
%  we show that many of the remaining errors are unlikely to be fixed through architecture manipulation.  Instead, more effort in the future should be paid on exploring resources.
\end{abstract}

\begin{CJK}{UTF8}{gbsn}

\section{Introduction}

Neural networks have become ubiquitous in natural language processing.  
For the word segmentation task, there has been a growing body of work exploring novel neural network architectures for learning useful representation and thus better segmentation prediction \cite{P14-1028,ma-hinrichs:2015:ACL-IJCNLP,DBLP:conf/acl/ZhangZF16, DBLP:journals/corr/LiuCGQL16, cai2017fast, DBLP:journals/corr/abs-1711-04411}.

We show that properly training and tuning a relatively simple architecture with a minimal feature set and greedy search achieves state-of-the-art accuracies and beats more complex neural-network architectures.
Specifically, the model itself is a straightforward stacked bidirectional LSTM (Figure \ref{fig_models}) with just two input features at each position (character and bigram).
We use three widely recognized techniques to get the most performance out of the model: pre-trained embeddings \cite{yang2017neural,zhou2017word}, dropout \cite{srivastava2014dropout}, and hyperparameter tuning \cite{weiss2015structured,melis2018on}.
These results have important ramifications for further model development. Unless best practices are followed, it is difficult to compare the impact of modeling decisions, as differences between models are masked by choice of hyperparameters or initialization.

In addition to the simpler model we present, we also aim to provide useful guidance for future research by examining the errors that the model makes.
About a third of the errors are due to annotation inconsistency, and these can only be eliminated with manual annotation.
The other two thirds are those due to out-of-vocabulary words and those requiring semantic clues not present in the training data.
Some of these errors will be almost impossible to solve with different model architectures.
For example, while 抽象概念 (abstract concept) appears as one word at test time, any model trained only on the MSR dataset will segment it as two words: 抽象 (abstract) and 概念 (concept), which are seen in the training set 28 and 90 times, respectively, and never together.
Thus, we expect that iterating on model architectures will give diminishing returns, while leveraging external resources such as unlabeled data or lexicons is a more promising direction.

%Predictions on these out-of-vocabulary (OOV) words are helped by pretraining character and bigram embeddings, but they remain a significant source of errors.
%There is also a small amount of errors being ambiguities that require semantic clue, which in turn requires training signal that beyond the labeled training set.
%Thus, our observation suggests that instead of pursing advanced neural-network architectures,  leveraging external resources (e.g., unlabeled data) is a more promising direction.  
%A significant portion are due to annotation inconsistencies in the data, where the same words are segmented differently at different locations in the data.
%Another significant portion of errors is due to over- or under- segmentation.

In sum, this work contributes two significant pieces of evidence to guide further development in Chinese word segmentation. First, comparing different model architectures requires careful tuning and application of best practices in order to obtain rigorous comparisons. Second, iterating on neural architectures may be insufficient to solve the remaining classes of segmentation errors without further efforts in data collection.

\section{Model}

\begin{figure}
\small
\centering
\scalebox{0.8}{
\begin{tikzpicture}[node distance=0.9cm,bend angle=45,auto]
  \tikzstyle{char}=[circle,thick,draw=blue!75,fill=blue!20,minimum size=4mm]
  \tikzstyle{LSTM}=[FF,draw=red!75,fill=red!20]
  \tikzstyle{FF}=[rectangle,thick,draw=black!75, fill=black!20,minimum size=4mm]
  \begin{scope}
    \node[](char)[text width=1.2cm] {characters};
    \node[](lstmb)[above of=char,text width=1.2cm] {backward LSTM};
    \node[](lstm)[above of=lstmb,text width=1.2cm] {forward LSTM};
    \node[](pred)[above of=lstm,text width=1.2cm] {softmax};
    \foreach \x in {0,...,2}
    {
      \node[LSTM,xshift=\x * 1cm + 0.5cm](lb\x)[right of=lstmb] {\ };
      \node[LSTM,xshift=\x * 1cm + 0.5cm](l\x)[right of=lstm] {\ };
      \node[char,xshift=\x * 1cm + 0.5cm](c\x)[right of=char] {};
      \node[xshift=\x * 1cm + 0.5cm](p\x)[right of=pred] {BIES};
      \draw[->,thick] (c\x) edge (lb\x);
      \draw[->,thick] (l\x) edge (p\x);
      \draw[->, thick, bend right] (lb\x) edge (p\x);
      \draw[->,thick, bend left] (c\x) edge (l\x);
    }
    \draw[->,thick] (lb2) edge (lb1) (lb1) edge (lb0);
    \draw[<-,thick] (l1) edge (l0) (l2) edge (l1);
    \node[yshift=0.3cm](a)[below of=c1] {(a)};
  \end{scope}

  \begin{scope}[shift={(5cm,0)}]
    \node[](char)[text width=1.2cm] {characters};
    \node[](lstmb)[above of=char,text width=1.2cm] {backward LSTM};
    \node[](lstm)[above of=lstmb,text width=1.2cm] {forward LSTM};
    \node[](pred)[above of=lstm,text width=1.2cm] {softmax};
    \foreach \x in {0,...,2}
    {
      \node[LSTM,xshift=\x * 1cm + 0.5cm](lb\x)[right of=lstmb] {\ };
      \node[LSTM,xshift=\x * 1cm + 0.5cm](l\x)[right of=lstm] {\ };
      \node[char,xshift=\x * 1cm + 0.5cm](c\x)[right of=char] {};
      \node[xshift=\x * 1cm + 0.5cm](p\x)[right of=pred] {BIES};
      \draw[->,thick] (c\x) edge (lb\x) (lb\x) edge (l\x) (l\x) edge (p\x); %(ff\x) (ff\x) edge (p\x);
    }
    \draw[->,thick] (lb1) edge (lb0) (lb2) edge (lb1);
    \draw[<-,thick] (l1) edge (l0) (l2) edge (l1);
    \node[yshift=0.3cm](b)[below of=c1] {(b)};
  \end{scope}
\end{tikzpicture}
}
\vspace*{-2.5em}
\caption{
\label{fig_models}
Bi-LSTM models: (a) non-stacking, (b) stacking.  Blue circles are input (char and char bigram) embeddings.  Red squares are LSTM cells.  BIES is a 4-way softmax.
}
\end{figure}
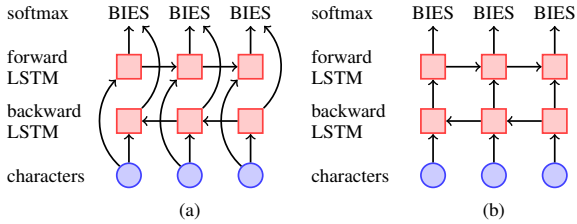

Our model is relatively simple. Our approach uses long short-term memory neural networks architectures (LSTM) since previous work has found success with these models \cite[][inter alia]{chen2015long,zhou2017word}.
We use two features: unigrams and bi-grams of characters at each position. These features are embedded, concatenated, and fed into a stacked bidirectional LSTM (see Figure~\ref{fig_models}) with two total layers of 256 hidden units each.
The softmax layer of the bi-LSTM predicts Begin/Inside/End/Single tags encoding the relationship from characters to segmented words. 

In the next sections we describe the best practices we used to achieve state-of-the-art performance from this architecture. Note that all of these practices and techniques are derived from related work, which we describe.

\paragraph{Recurrent Dropout.}

Contrary to the recommendation of \newcite{zaremba2014recurrent}, we apply dropout to the recurrent connections of our LSTMs, and we see similar improvements when following the recipe of \newcite{gal2015theoretically} or simply sample a new dropout mask at every recurrent connection. 

\paragraph{Hyperparameters.}

We use the momentum-based averaged SGD procedure from \cite{weiss2015structured} to train the model, with few additions.
We normalized each gradient to be at most unit norm, and used asynchronous SGD updates to speed up training time.
For each configuration we evaluated, we trained different settings of a manually tuned hyperparameter grid, varying the initial learning rate, learning rate schedule, and input and recurrent dropout rates. 
We fixed the momentum parameter $\mu = 0.95$. 
The full list of hyperparameters is given in Table~\ref{tab_grids}.
We show the impact of this tuning procedure in Table~\ref{tab_hyperparameters}, which we found was crucial to measure the best performance of the simple architecture. 

\paragraph{Pretrained Embeddings.} Pre-training embedding matrices
from automatically gathered data is a powerful technique that has been
applied to many NLP problems for several years
(e.g. \citet{collobert2011natural,mikolov2013efficient}).  We pretrain
the character embeddings and character-bigram embeddings using
wang2vec\footnote{\url{https://github.com/wlin12/wang2vec}}
\cite{ling2015two}, which modifies word2vec by incorporating
character/bigram order information during training.
Note that this idea has been used in segmentation previously by \newcite{zhou2017word}, but they also augment the contexts by adding the predictions of a baseline segmenter as an additional context.
We experimented with both treating the pretrained embeddings as constants or fine-tuning on the particular datasets.
\begin{table}
  \centering
  \scalebox{0.92}{
  \begin{tabular}{l|rrr}
   &{} Train & Development &  Test \\
   \hline
    AS    & 4,903,564  & 546,017 &  122,610 \\                 
    CTIYU & 1,309,208  & 146,422 &  40,936 \\
    MSR   & 2,132,480  & 235,911 &  106,873 \\
    CTB6  & 641,368    & 59,954  &  81,578  \\
    CTB7  & 950,138    & 59,954  &  81,578 \\
    PKU   & 994,822    & 115,125 &  104,372 \\
    UD    & 98,608     & 12,663  &  12,012 \\
   \hline
      \end{tabular}
  }
  \caption{
    \label{statistics}
    Statistics of training, development and test set.
  }
\end{table}

\begin{table*}
  \centering
  \scalebox{0.92}{
  \begin{tabular}{l|ccccccc}
   &{} AS & CITYU &   CTB6 & CTB7    &   MSR  &    PKU  & UD \\
   \hline
    \newcite{DBLP:journals/corr/LiuCGQL16}   & --- & --- & 95.9  &  ---  & 97.3 & \bf96.8  \\                 
    \newcite{yang2017neural}     & 95.7    & 96.9 & 96.2 &  ---  & 97.5 & 96.3   & --- \\
    \newcite{zhou2017word}       & ---     & ---  & 96.2 &  ---  & 97.8 & 96.0   & --- \\
    \newcite{cai2017fast}        & ---     & 95.6 & ---  &  ---  & 97.1 & 95.8   & --- \\
    \newcite{kurita2017neural}   & ---     & ---  & ---  &  96.2 & ---  & ---    & --- \\
    \newcite{chen2017adversarial}& 94.6    & 95.6 & 96.2 &  ---  & 96.0 & 94.3   & --- \\
    \newcite{K17-3015}           & ---     & ---  & ---  &  ---  & ---  & ---    & 94.6 \\
    \newcite{DBLP:journals/corr/abs-1711-04411}   & ---     & ---  & ---  &  ---  & 98.0  & 96.5  & --- \\
   \hline
     Ours (fix embedding)        & \bf96.2 & \bf97.2 & \bf96.7 &  \bf96.6 &  97.4  & 96.1    & \bf96.9 \\
     Ours (update embedding)     & 96.0 &  96.8   & 96.3    &  96.0 &  \bf98.1  & 96.1    & 96.0 \\
%    \hline
%    \newcite{chen2015long}$\dagger$  & ---     & ---  & 96.0 &  ---  & 97.4 & 96.5 & --- \\
  \end{tabular}
  }
  \caption{
    \label{tab_state_of_the_art}
    The state of the art performance on different datasets. For \newcite{kurita2017neural} and  \newcite{chen2017adversarial} we report their best systems (segpos+dep and Model-I-ADV respectively).
    $\dagger$Not directly comparable to the rest of the table due to the usage of an external dictionary. Our bolded results are significantly better (p $<$ 0.05 bootstrap resampling) except on MSR.
  }
\end{table*}

\begin{table*}
  \centering
  \scalebox{0.92}{
  \begin{tabular}{l|cccccccc}
    &{} AS & CITYU &   CTB6 & CTB7    &   MSR  &    PKU  & UD \\
   \hline
    \newcite{DBLP:journals/corr/LiuCGQL16}   & --- & --- & 94.6  &  ---  & 94.8 & 94.9   & --- \\
    \newcite{zhou2017word}   & ---   & ---  & 94.9 &  ---  & 97.2 & 95.0   & --- \\
    \newcite{cai2017fast}    & 95.2  & 95.4 & ---  &  ---  & 97.0 & \bf95.4   & --- \\
    \newcite{DBLP:journals/corr/abs-1711-04411}   & ---     & ---  & ---  &  ---  & 96.7  & 94.7    & --- \\
   \hline
     Ours     & \bf95.5  & \bf95.7 & \bf95.5 & \bf95.6 & \bf97.5 & \bf95.4 & \bf94.6 \\
  \end{tabular}
  }
  \caption{
    \label{close_test}
    Performance of recent neural network based models without using pretrained embeddings.  Our model's wins are statsitically significantly better than prior work (p $<$ 0.05 bootstrap resampling), except on PKU.
  }
\end{table*}

\paragraph{Other Related Work.}
Recently, a number of different neural network based models have been proposed for word segmentation task.  One common approach is to learn word representation through the characters of that word.  For example, \newcite{DBLP:journals/corr/LiuCGQL16} runs bi-directional LSTM over characters of the word candidate and then concatenate bi-directional LSTM outputs at both end points. \newcite{cai2017fast} adopts a gating mechanism to control relative importance of each character in the word candidate. 

Besides modeling word representation directly, sequential labeling is another popular approach.  For instance, \newcite{D13-1061} and \newcite{P14-1028} predict the label of a character based context of a fixed sized local window. \newcite{chen2015long} extends the approach by using LSTMs to capture potential long distance information.  Both \newcite{chen2015long} and \newcite{P14-1028} use a transition matrix to model interaction between adjacent tags. \newcite{zhou2017word} conduct rigorous comparison and show that such transition matrix rarely improves accuracy.  Our model is similar to \newcite{zhou2017word}, except that we stack the backward LSTM on top of the forward one, which improves accuracy as shown in later section.

Our model is also trained via a simple maximum likelihood objective. In contrast, other state-of-the-art models use a non-greedy approach to training and inference, e.g. \newcite{yang2017neural} and \citet{zhang2016transition}.

\begin{table*}
\centering
\begin{tabular}{lr}
\toprule
Parameter & Values \\ \hline
Char embedding size & [64] \\
Bigram embedding size & [16, 32, 64] \\
Learning rate & [0.04, 0.035, 0.03] \\
Decay steps & [32K, 48K, 64K] \\
Input dropout rate & [0.15, 0.2, 0.25, 0.3, 0.4, 0.5, 0.6] \\ 
LSTM dropout rate & [0.1, 0.2, 0.3, 0.4] \\
\bottomrule
\label{tab_grids}
\end{tabular}
\caption{Hyperparameter settings. }
\end{table*}

\begin{table*}
\centering
\begin{tabular}{lrrrrrrr}
  &  AS & CITYU & CTB6 & CTB7 & MSR & PKU & UD \\ \hline
OOV \% & 4.2 & 7.5 & 5.6 & 5.0 & 2.7 &  3.6 & 12.4  \\ \hline
Recall \% (random embedding) & 65.7 & 75.1 & 73.4 & 74.1 & 71.0 & 66.0 & 81.1 \\
Recall \% (pretrain embedding) & 70.7 & 87.5 & 85.4 &  85.6 & 80.0 & 78.8 & 89.7 \\ 
\end{tabular}
\caption{
\label{test_set_oov} Test set OOV rate, together with OOV recall achieved with randomly initialized and pretrained embeddings, respectively.
}
\end{table*}

\begin{table*}
\centering
\scalebox{0.92}{
\begin{tabular}{r|rrrrrrr|r}
System &  AS & CITYU & CTB6 & CTB7 & MSR & PKU & UD  & Average \\
\hline
This work  &  98.03 & 98.22 & 97.06 & 97.07 & 98.48 & 97.95 & 97.00  & 97.69  \\
-LSTM dropout  &  +0.03 & -0.33 & -0.31 & -0.24 & +0.04 & -0.29 & -0.76  & -0.35  \\
-stacked bi-LSTM  &  -0.13 & -0.20 & -0.15 & -0.14 & -0.17 & -0.17 & -0.39  & -0.27  \\
-pretrain  &  -0.13 & -0.23 & -0.94 & -0.74 & -0.45 & -0.27 & -2.73  & -0.78  \\
\end{tabular}
}
\caption{
\label{tab_ablation}
Ablation results on development data.  Top row: absolute performance of our system. Other rows: difference relative to the top row.
}
\end{table*}
\section{Experiments}

\paragraph{Data.} We conduct experiments on the following datasets:  
Chinese Penn Treebank 6.0 (CTB6) with data split according the official document; 
Chinese Penn Treebank 7.0 (CTB7) with recommended data split \cite{wang-EtAl:2011:IJCNLP-2011};
Chinese Universal Treebank (UD) from the Conll2017 shared task \cite{zeman-EtAl:2017:K17-3} with the official data split;
Dataset from SIGHAN 2005 bake-off task (Emerson, 2005).  
Table~\ref{statistics} shows statistics of each data set.
For each of the SIGHAN 2005 dataset, we randomly select $10\%$ training data as development set.  We convert all digits, punctuation and Latin letters to half-width, to handle full/half-width mismatch between training and test set.
We train and evaluate a model for each of the dataset, rather than train one model on the union of all dataset. Following \newcite{yang2017neural}, we convert AS and CITYU to simplified Chinese. 

%\paragraph{Hyperparameters.} Table~\ref{tab_hyperparameter} lists hyperparameters 

\subsection{Main Results}
Table~\ref{tab_state_of_the_art} contains the state-of-the-art results from recent neural network based models, together with the performance of our model.
Table ~\ref{close_test} contains results achieved without using any pretrained embeddings. 

Our model achieves the best results among NN models on 6/7 datasets.
In addition, while the majority of datasets work the best if the pretrained embedding matrix is treated as constant, the MSR dataset is an outlier: fine-tuning embeddings yields a very large improvement.
We observe that the likely cause is a low OOV rate in the MSR evaluation set compared to other datasets.

\subsection{Ablation Experiments}
\label{sec_ablation}

To see which decisions had the greatest impact on the result, we
performed ablation experiments on the holdout sets of the different
corpora.  Starting with our proposed system\footnote{Based on development set accuracy, we keep the pretrained embedding fixed for all datasets except MSR and AS. }, we remove one
decision, perform hyperparameter tuning, and see the change in
performance. The results are summarized in Table~\ref{tab_ablation}.
Negative numbers in Table~\ref{tab_ablation} correspond to decreases in performance for the ablated system.  Note that although each of the components help performance on average, there are cases where we observe no impact.  For example using recurrent dropout on AS and MSR rarely affects accuracy.
%For example, pretrained embeddings actually hurt performance for the MSR data set.  This is because the out of vocabulary rate is very low, and we use fixed pretrained embeddings.  If we allow the model to update the embeddings, pretrained embeddings improve performance on MSR as well.

We next investigate how important the hyperparameter tuning is to this
ablation.  In the main result, we tuned each model separately for each
dataset.  What if instead, each model used a single hyperparameter
configuration for all datasets?  In Table \ref{tab_hyperparameters}, we
compare fully tuned models with those that share hyperparameter
configurations across dataset for three settings of the model.  We can 
see that hyperparameter tuning consistently improves model accuracy across all settings.
%We find that tuning hyperparameters account for significant portion of
%the results (roughly 0.5\% improvement) and that, in particular,
%using the stacked hyperparameters for non-stacked architectures yields
%a significant drop in accuracy (roughly 1 \%).

% We find that 
% approach to a Here, we compute the optimal global hyperparameter setting from the grid across all datasets for each model (``Best''). We also investigate whether or not the hyperparameters from one model generalize to the others: for each model, we apply the best configuration from the others and compute the {\em lowest} score (``Worst''). We only apply this to models where we tuned recurrent dropout rates which have equivalent hyperparameter grids.

% This suggests that in order to properly compare between different architectures, the hyperparameters for each must be carefully tuned.
% Our results also suggest that when comparing to prior work, it's important to understand if the hyperparameters were tuned for the particular datasets reported on, and perform proper per-dataset tuning if so. 

\begin{table}
\centering
\begin{tabular}{l|ccc}
  System & Fully tuned & Avg \\% & Mismatch \\
  \hline
  This work & 97.69 & 97.49 \\%& 96.84\\
  -Stacked  & 97.41 & 97.16 \\%& 95.91 \\
  -Pretraining  & 96.90 & 96.81 \\%& 96.11 \\
\end{tabular}
\caption{
\label{tab_hyperparameters}
Hyperparameter ablation experiments. ``Fully tuned'' indicates per-system tuning for each dataset. ``Avg'' is the best setting when averaging across datasets. %To see how one model's hyperparameters work for another model, ``Mismatch'' takes the three ``Avg'' hyperparameter settings, applies them to the others, and computes the minimum accuracy.
}
\end{table}

\subsection{Error Analysis}

In order to guide future research on Chinese word segmentaion, it is 
important to understand the types of errors that the system is making.
To get a sense of this, we randomly selected 54 and 50 errors from the CTB-6 and MSR test set, respectively.
We then manually analyzed them. 
%Table ~\ref{tab_over_under} summarizes the statistics of each error type as well as errors caused by annotation inconsistency.

%todo(maji): update statistics.
%\begin{table}
%\begin{tabular}{c|c|c}
%Error type & Count & Fraction \\
%\hline
%Overlapping & 2& 4\% \\ 
%Undersegmentation & 11  & 20\% \\
%Oversegmentation & 13  & 24\% \\
%Inconsistent annotation & 28 & 52\% \\
%\end{tabular}
%\caption{
%\label{tab_over_under}
%Error type distribution for the model proposed in this work on the CTB-6 test data.
%}
%\end{table}

%{\bf Under-segmentation} denotes cases where the several gold words are merged into one token. 
The model learns to remember words it has seen, especially for high frequency words.  It also learns the notion of prefixes/suffixes, which aids predicting OOV words, a major source of segmentation errors \cite{HUANGChang-ning:8}. 
Using pretrained embeddings enables the model to expand the set of prefixes/suffixes through their nearest neighbors in the embedding spaces, and therefore further improve OOV recall (on average, using pretrained embeddings contributes to $10\%$ OOV recall improvement, also see Table~\ref{test_set_oov} for more details).

Nevertheless, OOV remains challenging especially for those that can be divided into words frequently seen in the training data, and most (37 out of 43) of the oversegmentation errors are due to this.
For instance, the model incorrectly segmented the OOV word 抽象概念 (abstract concept) as 抽象 (abstract) and 概念 (concept).  抽象 and 概念 are seen in the training set for 28 times and 90 times, respectively.   
Unless high coverage dictionaries are used, it is difficult for any supervised model to learn not to follow this trend in the training data.

In addition, the model sometimes struggles when a prefix/suffix can also be a word by itself.
For instance, \hichar{权} (right/power) frequently serves as a suffix, such as 管理\hichar{权} (right of management), 立法\hichar{权} (right of legislation) and 终审\hichar{权} (right of final judgment). 
When the model encounters 下放 (delegate/transfer)  \hichar{权}(power), it incorrectly merges them together.

Similarly, the model segments \hichar{居} (in/at) + 中 (middle) as \hichar{居}中 (in the middle), since the training data contains words such as \hichar{居}首 (in the first place) and \hichar{居}次 (in the second place). This example also hints at the ambiguity of word delineation in Chinese, and explains the difficulty in keeping annotations consistent.  

As another example, \hichar{县} is often attached to another proper noun to become a new word, e.g., 高雄 (Kaohsiung) + \hichar{县} becomes  高雄县 (county of Kaohsiung), 新竹(Hsinchu) + \hichar{县} becomes  新竹\hichar{县} (county of Hsinchu).  When seeing
银行\hichar{县}支行 (bank's county branch), which should be 
银行 (bank) + \hichar{县}支行 (county branch), the model outputs 银行\hichar{县} + 支行 (i.e. a county named bank).
Fixing the above errors requires semantic level knowledge such as `Bank' (银行) is unlikely to be the name of a county (\hichar{县}), and likewise, transfer power (下放\hichar{权}) is not a type of right (\hichar{权}).

Previous work \cite{HUANGChang-ning:8} also pointed out that OOV is a major obstacle to achieving high segmentation accuracy.  They also mentioned that machine learning approaches together with character-based features are more promising in solving OOV problem than rule based methods.  Our analysis indicate that learning from the training corpus alone can hardly solve the above mentioned errors. Exploring other sources of knowledge is essential for further improvement.
One potential way to acquire such knowledge is to use a language model that is trained on a large scale corpus \cite{DBLP:journals/corr/abs-1802-05365}. We leave this to future investigation. 

Unfortunately, a third (34 out of 104) of the errors we have looked at were due to annotation inconsistency.  
For example, 建筑系 (Department of Architecture) is once annotated as 建筑 (Architecture) + 系 (Department) and once as 建筑系  under exactly the same context 建筑系教授喻肇青 (Zhaoqing Yu, professor of Architecture).
高新技术 (advanced technology) is annotated as 高 (advanced) + 新 (new) + 技术 (technology) for 37 times, and is annotated as 高新 (advanced and new) + 技术 (technology) for 19 times.

In order to augment the manual verification we performed above, we also wrote a script to automatically find inconsistent annotations in the data.  Since this is an automatic script, it cannot distinguish between genuine ambiguity and inconsistent annotations.  The heuristic we use is the following: for all word bigrams in the training data, we see if they also occur as single words or word trigrams.  We ignore the dominant analysis and count the number of occurrences of the less frequent analyses and report this number as a fraction of the number of tokens in the corpus.
Table~\ref{tab_auto_inconsistency} shows the results of running the script.  We see that the AS corpus is the least consistent (according to this heuristic) while MSR is the most consistent.  This might explain why both our system and prior work have relatively low performance on AS even though this has the largest training set.  By contrast results are much stronger on MSR, and this might be in part because it is more consistently annotated.
The ordering of corpora by inconsistency roughly mirrors their ordering by accuracy.
  
%, such as annotation inconsistency, OOV and segmentation ambiguities.  They also pointed out that   

\begin{table}
\centering
\begin{tabular}{lrr}
\toprule
{} &   tokens &  inconsistency \% \\
corpus &          &                  \\
\midrule
AS     &  4,903,564 &    1.31 \\
CITYU  &  1,309,208 &    0.62 \\
CTB6   &   641,368 &     1.27 \\
CTB7   &   950,138 &         1.64 \\
MSR    &  2,132,480 &         0.28 \\
PKU    &   994,822 &         0.53 \\
UD     &    98,608 &         0.46 \\
\bottomrule
\end{tabular}
\caption{
\label{tab_auto_inconsistency} Automatically computed inconsistency in the corpus training data.  See text for methodology.
}
\end{table}

\end{CJK}

\section{Conclusion}

In this work, we showed that further research in Chinese segmentation must overcome two key challenges: (1) rigorous tuning and testing of deep learning architectures and (2) more effort should be made on exploring resources for further performance gain. %In future work, we will explore new data collection tasks that can to alleviate these challenges.

\newpage 
}

\bibliography{emnlp2018}
\bibliographystyle{acl_natbib_nourl}

\clearpage

\thesupplementarycontent{

\appendix

\section{Supplemental Material}
\label{sec:supplemental}

}

\end{document}